\def\year{2019}\relax
\documentclass[letterpaper]{article}

\usepackage{times}               % Required
\usepackage{helvet}              % Required
\usepackage{courier}             % Required
\usepackage{url}                 % Required
\usepackage{graphicx}            % Required
\usepackage{aaai19}              % Required
\usepackage[titlenumbered,ruled]{algorithm2e}
\newcommand{\citeinline}[1]{\citeauthor{#1}~(\citeyear{#1})}
\usepackage[utf8]{inputenc}
\usepackage{amsmath}
\usepackage{amssymb}
\usepackage{mathtools}
\usepackage{parskip}
\usepackage{tabularx}
\usepackage{subfigure}
\usepackage[disable]{todonotes}

\DeclareMathOperator*{\argmin}{arg\,min}
\newcommand{\tdef}{\Theta_{def}}

\title{Learning Multiple Defaults for Machine Learning Algorithms}
\author{Florian Pfisterer$^\ast$ \ and Jan N. van Rijn$^\dag$ \ and  Philipp Probst$^\ast$ \ and Andreas M\"{u}ller$^\dag$ \ and Bernd Bischl$^\ast$\\
$^\ast$Ludwig Maximilian University of Munich, Germany\\
$^\dag$Columbia University, New York, U.S.A.}
\date{September 2018}

\begin{document}
\maketitle

\begin{abstract}
The performance of modern machine learning methods highly depends on their \textit{hyperparameter configurations}.
One simple way of selecting a configuration is to use \textit{default settings}, often proposed along with the publication and implementation of a new algorithm.
Those default values are usually chosen in an ad-hoc manner
to work \textit{good enough} on a wide variety of datasets.
To address this problem, different automatic hyperparameter configuration algorithms have been proposed, which select an optimal configuration per dataset. 
This principled approach usually improves performance, but adds additional algorithmic complexity 
and computational costs to the training procedure.
As an alternative to this, we propose learning a set of complementary default values from a large database of prior empirical results. 
Selecting an appropriate configuration on a new dataset then requires only a simple, efficient and embarrassingly parallel search over this set.
We demonstrate the effectiveness and efficiency of the approach we propose in comparison to random search and Bayesian Optimization.
\end{abstract}

\section{Introduction}

%%%% Background information (what is known) %%%
The performance of most machine learning algorithms highly depends on their hyperparameter settings. 
Various methods exist to automatically optimize hyperparameters, including random search~\cite{Bergstra2012}, Bayesian optimization~\cite{Snoek2012,Hutter2011}, meta-learning~\cite{Brazdil2008} and bandit-based methods~\cite{Li2017}.
Depending on the algorithm, properly tuning the hyperparameters yields a considerable performance gain~\cite{Lavesson2006}. 

%%% Knowledge gap (what is not yet known) %%%
Despite this acknowledged importance of tuning the hyperparameters, in many practical cases it is often neglected.

Possible reasons for this are the additional run time, code complexity and experimental design questions.
It has indeed been pointed out that properly deploying a hyperparameter tuning strategy requires expert knowledge~\cite{Probst2018,Rijn2018}.

When parameters are not tuned, they are often set to a default value provided by the software authors. While not tuning parameters at all can be detrimental, defaults provide a fall-back for cases, where no additional knowledge is available.
\citeinline{wistuba15b} proposed to extend the notion of pre-specified defaults to ordered sets of defaults, combining the prior knowledge encoded in default values with the flexibility of optimization procedures. This work directly builds upon this notion. 
Our ordered sets of defaults are diverse lists of parameter settings for a particular algorithm, ordered by their performance across datasets.
This can be seen as an extension of the classical exhaustive grid-search: Instead of searching all possible combinations in the grid, we keep only those configurations that historically (on a collection of benchmark datasets) performed well. Given that we eliminate most candidates using prior data, we can then afford to start with a very fine grid, approximating the results of a continuous optimization procedure.

A different perspective on multiple defaults is as a special case of meta-learning: we build a model using a collection of benchmark datasets, that allows us to predict good candidate parameters for a new dataset. Only we do not use any properties of the new dataset, and always predict the same ordered set of candidates.

Compared with more complex optimization procedures, multiple defaults have several benefits.

\textbf{Ease of implementation}
    Sets of defaults can be easily computed in advance and implemented as look-up tables. Only simple resampling is required to select the best configuration from the set.
    
\textbf{Ease of use}
    The concept of multiple defaults is easy to understand and does not introduce any additional parameters or specification of parameter ranges. The number of default configurations to be evaluated is determined by computational constraints.
    
\textbf{Strong anytime performance}
    Defaults can achieve high performances even if only few evaluations can be performed. If additional computational resources are available, they 
    can be used in combination with other optimization methods, e.g., as good initial values for conventional tuning methods.
    
\textbf{Embarrassingly parallel}
    Evaluation of the ordered set of defaults can be arbitrarily parallelized.
    
\textbf{Robustness}
    Defaults do not suffer from the problems associated with optimizing over high-dimensional, mixed-type input spaces nor from potential crashes or failures of the optimization method.

We conjecture that a small set of well-performing configurations can perform quite well on a broad set of datasets. We will leverage a large set of historic datasets, and the performance results of prior experiments that are readily available on OpenML~\cite{Vanschoren2014,Rijn2016}.
While proper hyperparameter tuning techniques remain preferable when the resources and expertise are available, simply iterating over an orderd set of defaults might be a viable alternative when this is not the case. 

%%% Contributions %%%%
We define a model-agnostic approach of learning not only a single configuration parameter configuration, but instead a set of configurations, that \textit{together} perform well on new datasets.
That is, any given set of defaults should contain at least one parameter configuration that performs well on a given dataset.
These defaults can be written down and hard coded into software implementations, and thus easily be adapted by users.

\textbf{Our contributions} are the following. 
1) We describe two methods, an exact exhaustive method and a greedy method, that acquire a list of defaults, based on predictions from surrogate models. In particular, the surrogate models allow to scale the method in a realistic setting to arbitrary algorithms and sizes of hyperparameter spaces.
2) we show that solving the underlying problem in an exact manner is NP-hard, 
3) due to this NP-hardness, we conduct a small experiment comparing the greedy and exact exhaustive approach, and 
4) empirically evaluate the defaults obtained through the greedy approach in a large benchmark using  6 configurable state-of-the-art ML algorithms, containing $2$--$10$ hyperparameters, on a wide range of datasets. 
\todo{I think we can make the argument that because greedy random search on surrogates is as good as exhaustive search, we will also be as good as greedy random search on discretized space?}
In this experiment we compare defaults found with the described method against random search as well as Bayesian Optimization. We show that the method we propose requires about $4$ times fewer model evaluations to achieve similar performances than random search or Bayesian optimization.

\section{Related work}
There are various openly available Machine Learning workbenches, implementing many algorithms. Some popular examples in scientific communities are  Weka~\cite{Hall2009}, scikit-learn~\cite{Pedregosa2011} and mlr~\cite{Bischl2016}.
Most algorithms have hyperparameters, that in turn have default values.
It has often been noted that using default values does not yield adequate performance, and can be improved by hyperparameter optimization~\cite{Bergstra2011,Bergstra2012,Hutter2011,Snoek2012,Li2017}. \citeinline{Lavesson2006} investigate for a given algorithm how strong the impact of different hyperparameter configurations can be across datasets. They infer that the importance of \textit{good} settings varies between algorithms, and that parameter tuning can be more important than the choice of an algorithm.
 
Many techniques have been proposed for performing hyperparameter optimization. 
\citeinline{Bergstra2012} compare grid search and random search, and concluded that although both approaches are rather simple, they already yield great performance gains compared to using a single default setting.
Furthermore, they noted that given the same number of iterations, random search is in many practical cases preferable over grid search. We will therefore use random search as the baseline in all our experiments.
Successive Halving~\cite{Jamieson2016} and Hyperband~\cite{Li2017} are full-exploration bandit-based methods, using initially small but increasing budgets to prioritize evaluation of particular hyperparameter settings.
The field of model based optimization (also often referred to as MBO or Bayesian Optimization) uses an internal \textit{empirical performance model}, which tries to learn a surrogate model of the objective function while optimizing it~\cite{Bergstra2011,Snoek2012,Hutter2011,mlrMBO}.
It focuses the search towards regions that are promising according to the model.

Alternatively, the field of meta-learning attempts to leverage knowledge obtained from experiments on prior datasets to a new dataset~\cite{Brazdil2008}. 
The underlying principle is to represent each dataset as a vector of numerical attributes. 
The assumption is that on datasets, similar algorithms that work well on these. 
A so-called meta-model can be trained to predict the performance of arbitrary configurations on new, unseen datasets~\cite{Gomes2012,Leite2012}.
Several approaches attempt to combine the paradigms of meta-learning and hyperparameter optimization, for example by warm starting hyperparameter optimization techniques~\cite{Feurer2015,Feurer2015b,Wistuba2015learning}, or in a streaming setting \cite{yogatama14}.
An approach very close to ours is investigated in \cite{wistuba15b}. From a perspective of finding good initial points for warm-stating Bayesian optimization, they propose to \textit{greedily} find a set of configurations, that minimizes the sum of risk across several datasets. This approach has severe limitations, we intend to alleviate in our work. First, the procedure requires hyperparameters evaluated on a grid across several datasets. This will scale exponentially with hyperparameter space dimensionality, and thus be practically infeasible for algorithms that require multiple hyperparameters. 
Additionally, it has been noted in \cite{Bergstra2012}, that grid search, especially when evaluated on coarse grid, often emphasizes regions that do not matter and suffers from poor coverage in important dimensions.

While all these methods yield convincing results and generated a considerable amount of scientific follow-up, they are by no means easy to deploy. 
Methods from the search paradigm require knowledge of which hyperparameters are important as well as suitable ranges to optimize over~\cite{Probst2018,Rijn2018}. 
Methods from the meta-learning paradigm require a set of historic datasets and meta-features to train on. 
Finding an informative set of training data and meta-features is still an open scientific question~\cite{Pinto2016,Rijn2016}. 

Most similar to the approach that we introduce are the works of \citeinline{Wistuba2015learning} and \citeinline{wistuba15b}. Additionally, the work of \citeinline{Feurer2018} also involves lists of defaults, although they do not detail on how to construct these. 

\citeinline{wistuba15b} propose a method that selects defaults based on meta-data from different algorithms. As the first reference to the multiple defaults, this work was a great contribution, but it also came with drawbacks. It requires a full grid of hyperparameters, evaluated on all tasks that have been encountered in the past. As this is not a realistic requirement, this makes applying this method unpractical, and it does not scale beyond a few hyperparameters. In fact, in the experimental evaluation only $2$ algorithms with no more than $4$ hyperparameters were considered. Parameters suggested by this method can only stem from the pre-specified grid.

Alternatively, \citeinline{Wistuba2015learning} propose a method that is able to warm-start Bayesian Optimization procedures, which can essentially also be seen as a set of defaults. 
The approach they propose requires a differentiable model, which in theory means that it can only work with numeric and unconditional hyperparameters. The method does not provide a fixed list of parameters, but instead adaptively learns them for a new problem instance.
The methods we propose do not require any of these. 
Additionally, the approach we propose does not rely on a predetermined budget of optimizations. For each different budget in terms of function evaluations, a different set of defaults will be recommended. 
It can therefore not successfully be applied in the more realistic case where the budget is run time.

Additionally, both works only experiment with $2$ algorithms and do not apply a \emph{nested} cross-validation procedure, which is required to make plausible conclusions when hyperparameter optimization is involved ~\cite{cawley10}.

\section{Method}
Consider a target variable $y$, a feature vector $X$, and an unknown joint distribution $P$ on $(X,y)$, from which we have sampled a dataset $\mathcal{D}$ containing $|\mathcal{D}|$ observations. 

A machine learning (ML) algorithm tries to approximate the functional relationship between $X$ and $y$ by producing a prediction model $\hat{f}_{\theta}(X)$, controlled by a multi-dimensional hyperparameter configuration $\theta \in \Theta$.
In order to measure prediction performance pointwise between a true label $y$ and its prediction $\hat{f}(X)$, we define a loss function $L(y, \hat{f}(X))$. 
We are naturally interested in estimating the expected risk of the inducing algorithm, w.r.t. $\theta$ on new data, also sampled from $\mathcal{P}$:
$R_{\mathcal{P}}(\theta) = E(L(y,\hat{f}(X))| \mathcal{P}).$
Thus, $R_{\mathcal{P}}(\theta)$ quantifies the expected predictive performance associated with a hyperparameter configuration $\theta$ for a given data distribution, learning algorithm and performance measure.

Given a certain data distribution, a certain learning algorithm and a certain performance measure, this mapping encodes the numerical quality for any hyperparameter configuration $\theta$.

Given $K$ different datasets (or data distributions) $\mathcal{P}_1,...,\mathcal{P}_K$, we arrive at $K$ hyperparameter risk mappings.
\[
R_k(\theta) = E(L(y,\hat{f}(X, \theta))| \mathcal{P}_k), \qquad k = 1,...,K.
\]

For a set of $M$ configurations $\Theta_M = \{\theta_1, \ldots, \theta_M\}$ 
and with a slight abuse of notation, we could define and visualize 
\[
R(\Theta_M) = \left( R_k(\theta_m) \right)_{k=1,\ldots,K; m=1,\ldots,M}
\]
as a matrix of dimensions $K \times M$ of risks for different configurations and datasets. 
Here, the $k$-th row-vector of $R(\Theta_M)$ contains the risks of all configurations in $\Theta_M$, evaluated on dataset $k$; while the $m$-column contains the empirical distribution of risk for $\theta_m$ across all datasets.

\subsection{Defining a set of optimal defaults}

Hyperparameter optimization methods usually try to find an optimal $\theta$ for a given dataset. In this work on the other hand, we try to find a fixed-size set $\Theta_{def}$ that works well over a wide variety of datasets, in the sense that $\Theta_{def}$ contains at least one configuration that works well on any given dataset (and in that case we do not really care about the performance of the other configurations on that dataset). 
In order for this to be feasible in practice, the individual datasets need to have at least some common structure from which we can generalize. This patterns can in general stem from algorithm properties, such as combinations of individual hyperparameters that work well together, or alternatively from similar data situations. By using a large number of datasets, we hope to find defaults that are less tailored to specific datasets, but generalize well. This allows focusing on the first kind of patterns. 
If patterns can be transferred from a set of datasets to a new dataset, one would assume that, given there exists a common structure, learned configurations perform significantly better than a set of randomly drawn data points on the same held out dataset. 

In practical terms, given our set $\Theta_{def}$, we would trivially evaluate all configurations in parallel (e.g., by cross-validation), and then simply select the best one to obtain the final fit for our ML algorithm.

\begin{algorithm}
\SetAlgoLined
\KwIn{Dataset $\mathcal{D}$, Inducer $\mathcal{A}$, candidate configurations $\tdef$ of size $n$}
\KwResult{Model induced by $\mathcal{A}$ on data $\mathcal{D}$}
 Cross-validate $\mathcal{A}$ on $\mathcal{D}$ with all $\theta_i \in \tdef$\;
 Select best $\theta^*$ from $\tdef$\;
 Fit $\mathcal{A}$ on complete $\mathcal{D}$ with $\theta^*$ and obtain ML model\;
 \caption{ML algorithm with multiple defaults}
 \label{alg:multdefs}
\end{algorithm}

Hence, an optimal set of defaults $\Theta_{def}$ should contain complementary defaults, i.e., some configurations $\theta \in \Theta_{def}$ can cover for shortcomings of other configurations from the same set.
This can be achieved by jointly optimizing over the complete set. 

\subsubsection{Risk of a set of configurations}
The risk of a set of configurations $\tdef \subset \Theta$, for datasets $1, \dots, K$ is given by:
\[
 % AM prefers option 2
G(\tdef) = h\left(\min_{j=1,\ldots,n} R_1(\theta_j),\dots,\min_{j=1,\ldots,n} R_K(\theta_j) )\right)
\]

We aggregate this to a single scalar performance value by using a function $h$, e.g., the median. For aggregation to be sensible, we assume that performances across all $K$ datasets are commensurable, which is a strong assumption.

The optimal set of defaults $\tdef$ of size $n$ is then given by 
\begin{align}\label{eq:main_objective}
\underset{\tdef \subset  \Theta, |\tdef| = n}{\argmin} \hspace{0.25cm}  G(\tdef)).
\end{align}

We compare two methods, namely an \textit{exact discretized} and a \textit{greedy} search approach, that allow us to obtain such sets of defaults.

\subsubsection{Computational Complexity} This problem is a generalization of the \emph{Maximum coverage problem}, which was proven NP-hard by~\citeinline{Nemhauser1978}. The original maximum coverage problem assumes Boolean input variables, s.t. each set covers covers certain elements, whereas the formulation in our context assumes scalar input variables. This adds additional complexity to the Exact Discretized Formulation. 

\subsection{Exact Discretized Optimization}
A discrete version of this problem can be formulated as a Mixed Integer Programming problem. 
The solution we propose is specific to the aggregation functions sum and mean. Other aggregation function can be incorporated as well, at the cost of introducing more variables and constraints. 
Given a discrete set of $M$ configurations $\{\theta_1, \ldots, \theta_M\} \subset \Theta$, we first define

\begin{equation}
    Q(k, m) = \{s : R_k(\theta_s) < R_k(\theta_m)\}
\end{equation}
for each $m \in \{1, \ldots, M\}$ and given $k$, where $R_k(\theta_s)$ is the empirical risk \todo{PP: now it is empirical?} of $\theta_s$ on $\mathcal{P}_k$.

Intuitively, $Q(k, m)$ now is a set of integer indices such that for each $q \in Q(k, m)$ holds that the risk of $\theta_q$ is lower than the risk of $\theta_m$ on dataset $k$.
The definition of $Q$ assumes no ties. 
Ties may be broken arbitrarily, but must be consistent.
For example, comparing $R_k(\theta_s) < R_k(\theta_m)$ can be replaced by the lexicographical comparison $(R_k(\theta_s), s) < (R_k(\theta_m), m)$.

In order to obtain a set of $n$ defaults, the goal is to minimize
\begin{equation}\label{eq:mip}
    \sum_{k=1}^{K} \sum_{m=1}^{M} \Psi_{k,m} \cdot R_k(\theta_m)
\end{equation}

subject to

\begin{equation}\label{eq:mip_numdefaults}
    \sum_{m=1}^{M} \phi_m = n
\end{equation}

\begin{equation}\label{eq:mip_penalize}
    \forall k: \forall m: \Psi_{k,m} \ge \phi_m - \sum_{s \in Q(k, m)} \phi_s
\end{equation}

\begin{equation}\label{eq:mip_prevent_negative}
    \forall k: \forall m: \Psi_{k,m} \ge 0
\end{equation}

\begin{equation}\label{eq:mip_aux}
    \forall k: \sum_{m=1}^{M} \Psi_{k,m} = 1 
\end{equation}
\todo{PP: m=1; maybe write the last expression first?}

The free variables are $\Psi$ (a matrix of size $K \times M$) and $\phi$ (a vector of size $M$, containing booleans). 
After the optimization procedure, $\phi_m = 1$ if and only if $\theta_m$ is part of the optimal set of defaults. Eq.~\ref{eq:mip_numdefaults} ensures that exactly the required number of defaults will be selected. 
Matrix $\Psi$ does not need to be restricted to a specific type, but will only contain values $\{0, 1\}$ (we will see why further on).
After the optimization procedure, element $\Psi_{k,m}$ will be $1$ if and only if configuration $\theta_m$ has the lowest risk on distribution $i$ out of all the configurations that are in the set of defaults. Formally, $\Psi_{k,m} = 1$ if and only if $\forall j: \phi_j = 0 \lor R_k(\theta_j) > R_k(\theta_i)$ (this is enforced by Eq.~\ref{eq:mip_penalize}).
The optimization criterion presented in Eq.~\ref{eq:mip} is the Hadamard product between the matrix $\Psi$ and the matrix of risks $R$. 
The outcome of this formula is equal to
the definition of the risk of a set of configurations, with $h$ being the sum. 
The aggregation functions sum and mean lead to the same set of defaults.

We will show that the constraints ensure the correct behaviour as described above. 
For an element $\Psi_{k,m}$, there are two factors that influence the minimum value: i) whether $\phi_i$ is part of the selected set of defaults, and ii) whether there are other configurations in the selected set of defaults that have a lower risk on distribution $k$.
If either or both conditions hold, $\Psi_{k,m} \ge 0$. If neither of the conditions hold, $\Psi_{k,m} \ge 1$.
Given that the risk is always positive, and matrix $\Psi$ can not contain negative numbers (Eq.~\ref{eq:mip_prevent_negative}), the optimizer will aim for as low as possible values in $\Psi$, in this case either $0$ or $1$.
The constraint presented in Eq.~\ref{eq:mip_aux} is formally not necessary, but removes the requirement that all values of $R$ need to be positive. 

\subsection{Greedy Search}\label{sec:greedy}

A computationally more feasible solution is to instead iteratively add defaults to the set in a greedy forward fashion, starting from the empty set as follows:

for i = $1, \dots , n$:\\
\begin{align}
\theta_{def, i} &:= \underset{\theta \in \Theta}{\argmin} \hspace{0.25cm}  G(\{ \theta \} \cup \Theta_{def, i-1}) \\
\Theta_{def, i} &:= \{ \theta_{def, 1}, \ldots , \theta_{def, i} \} 
\end{align}
where $\Theta_{def, i} = \emptyset$, and the final solution $\tdef = \Theta_{def, n}$.

An advantage of a greedy approach is that it results in a growing sequence of default subsets for increasing budgets. 
So if we compute a set of size $n=100$ through the above approach, e.g., in practice a user might opt to only evaluate the first $10$ configurations of the sequence (due to budget constraints) or to sequentially run through in them in an iterative process and stop when a desired performance is reached; in other words, it is an anytime algorithm. A possible disadvantage is, that for a given size $n$ the set of parameters might not be optimal according to Equation \ref{eq:main_objective}.

\subsection{Surrogate Models}
The exact discretized approach requires evaluating $R_k(\theta)$ on a fine discretization of the search space $\theta$, while greedy search even requires optimizing $G$, a complex function of $R_k(\theta)$. It is possible to estimate $R_k(\theta)$ empirically using cross-validation. In this case evaluation of $R_k(\theta)$ corresponds to evaluating a particular hyperparameter setting with cross-validation, which involves building several models.
While the proposed method only requires us to do this \emph{once} to obtain multiple defaults to be used in the future, building models on a fine grid is not tractable, especially when the number of hyperparameters is high.
Therefore we employ \emph{surrogate models} that predict the outcome of a given performance measure and algorithm for a given hyperparameter configuration. We train one model for each dataset on underlying performance data ~\cite{Eggensperger2015}.
This provides us with a fast approximate way to evaluate the performance of any given configuration, without the requirement of costly training and evaluating models using cross-validation.

Additionally, because we can not practically evaluate every $\theta \in \Theta$ on each dataset, as $|\Theta|$ can be infinitely big depending on the algorithm, we instead only evaluate a large random sample from $\Theta$. Cheap approximations can then be obtained via surrogate models.

\subsubsection{Standardizing results}
We mitigate the problem of lacking commensurability between datasets by normalizing performance results on a per-dataset basis to mean $0$ and standard deviation $1$ before training surrogate models. A drawback to this is, that some information with regards to the absolute performance of the algorithm and the spread across different configurations is lost.

\subsubsection{On the choice of an aggregation function}
Considering the fact that performances on different datasets are usually not commensurable \cite{Demsar2006}, an appropriate aggregation function or scaling is required to obtain sensible defaults.
One approach can be using quantiles. Depending on the choice of quantile, this either emphasizes low risks, i.e., datasets that are relatively easy anyways (by choosing the \textit{minimum}),  or high risks, i.e., hard datasets (when choosing the \textit{maximum}). This corresponds to optimizing an optimistic case (i.e., optimizing a best case scenario) or a pessimistic scenario (i.e., hedging against the worst case). Several other methods from decision theory literature, such as the Hodges-Lehmann criterion~\cite{hodges1952}  could also be used. From a theoretical point of view, it is not immediately clear which aggregation functions benefit the method most. From a small experiment, excluded for brevity, we concluded that in practice, the choice of an aggregation function has negligible impact on the performance of the set of defaults across datasets. Hence, we use the median over datasets as aggregation function.

\subsubsection{Defaults across algorithms}
In addition to providing a list of defaults for a given machine learning algorithm, the proposed method can also generate sets of defaults across a set of different learners. In this case, surrogate models need to be trained on results across all learners standardized per task. 
In this setting, the parameter space $\Theta$ is defined by a hyper-parameter that reflects the choice of an algorithm and the conjunction of the parameter spaces of all learners. 

\section{Experimental Setup}

We will perform two experiments.  We first present an experiment on small scale to compare the defaults obtained from the exact discretized approach against the greedy approach. As the exact discretized approach is presumably intractable, we can only perform a direct comparison on a small dataset, using a small number of defaults. 
In the second experiment, which is one of the main contributions of this work, we compare defaults obtained from the greedy approach against random search and Bayesian Optimization. This section describes the setup of these experiments. Due to the presumable intractability, the experiment comparing the greedy and exact discretized approach slightly deviates from this, in the sense that it operates on a discretized version of the problem, and uses a lower number of defaults (at most $6$). 

Estimations of the performance on future datasets can be obtained by evaluating a set of $n$ defaults $\Theta_{def}$ using Leave-One-Dataset-Out Cross-validation over $K$ datasets.
As a baseline, we compare to \textit{random search} with several budgets and \textit{Bayesian Optimization} with $32$ iterations. The latter simulates scenarios where the number of available evaluations is limited, for example due to computational constraints.

For each $k \in \{1, \dots , K\}$, we repeat the following steps:\\
\begin{itemize}
    \item \textbf{Defaults}\\for $n \in \{1, 2, 4, 8, 16, 32\}$:
    \begin{itemize}
        \item Learn a set of $n$ defaults $\Theta_{def}$ on datasets $\{ \{1, \dots , K\} \setminus k \}$
        \item Run the proposed \textit{greedy} algorithm with $\tdef$ on OpenML Task $k$ -- embedded in nested CV.
    \end{itemize}
    \item \textbf{Random search}\\for $i \in \{4, 8, 16, 32, 64\}$:
        \begin{itemize}
        \item Run random search with budget $i$ on OpenML Task $k$ -- embedded in nested CV.
        \end{itemize}
    \item \textbf{Bayesian Optimization}\\
    Run Bayesian Optimization budget $32$ on OpenML Task $k$ -- embedded in nested CV.
    %\item \textbf{Implementation Default}\\
    %Evaluate the implementation default on OpenML Task $k$ -- through normal CV. 
\end{itemize}

Performance estimates across all evaluated methods are obtained from a fixed outer 10-fold cross-validation loop on each left out dataset.
Evaluation of a set of configurations is done using \textit{nested} 5-fold cross-validation. 
For each configuration in the set, we obtain an estimation of the performance from the nested cross-validation loop.
This allows us to select a best configuration from the set, which is then evaluated on the outer test-set.
We use either \textit{mlrMBO} or \textit{scikit-optimize} as Bayesian Optimization frameworks.

\subsection{Datasets, Algorithms and Hyperparameters}
We use experimental results available on OpenML~\cite{Vanschoren2014,Rijn2016} to evaluate the sets of defaults. In total, we evaluate the proposed method on six algorithms, coming from \texttt{mlr}~\cite{Bischl2016} and \texttt{scikit-learn}~\cite{Pedregosa2011}. We use the $100$ datasets from the OpenML100~\cite{Bischl2017}. These contain between $500$ and $100,\!000$ observations, up to $5,\!000$ features and are not imbalanced. We evaluate the \texttt{mlr} algorithms on the 38 binary class datasets; we evaluate the \texttt{scikit-learn} algorithms on all 100 datasets. This decision was made based on the availability of meta-data. 

For \texttt{mlr}, we evaluate the method on \texttt{glmnet} \cite{glmnet} (elastic net implementation, 2 hyperparameters), \texttt{rpart} \cite{rpart} (decision tree implementation, 4 hyperparameters) and \texttt{xgboost} \cite{xgboost} (gradient boosting implementation, 10 hyperparameters). The optimization criterion was Area under the ROC curve.
We obtained in the order of a million results for randomly selected parameters on the $38$ binary datasets.

As of \texttt{scikit-learn}, we evaluate the method using Adaboost (5 hyperparameters), SVM (6 hyperparameters) and random forest (6 hyperparameters). The optimization criterion was predictive accuracy.  We obtained approximatelly $137.000$ results for randomly selected parameters of the 3 algorithms on the $100$ datasets. The hyperparameters and their respective ranges are the same as by~\citeinline{Rijn2018}.

All following results are obtained by computing defaults using Leave-One-Dataset-Out cross-validation. This means we iteratively learn defaults using $K - 1$ datasets, and evaluate using the held-out dataset.
Defaults thus have not been learned from the dataset that they are evaluated on. For any given configuration, we can either obtain an approximation of the risk, by resorting to trained surrogate models, or estimate the true performance using cross-validation.

\begin{figure}[tb!]
  \begin{center}
    \includegraphics[width=.965\columnwidth]{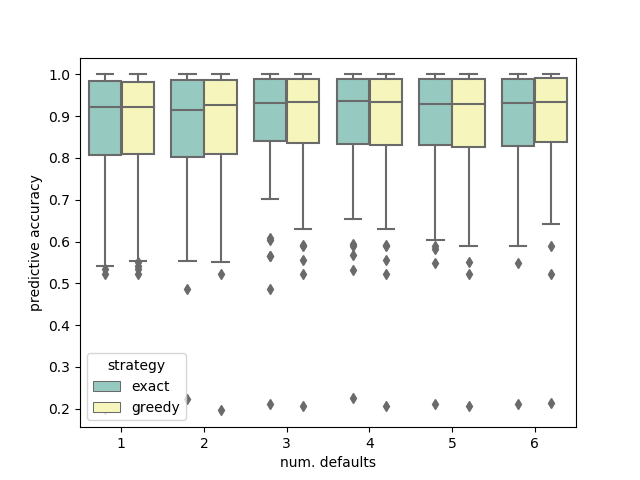}
    \caption{Performance of defaults obtained by exact discretized vs. the greedy. \label{fig:exact}}
  \end{center}
\end{figure}

 \begin{figure*}[tb!]
   \begin{center}
     \subfigure[Elastic net] {
       \includegraphics[width=.3\textwidth]{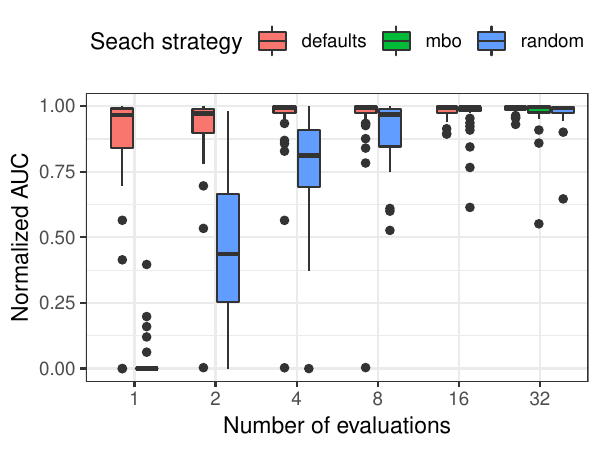}
     }
     \subfigure[Decision tree] {
       \includegraphics[width=.3\textwidth]{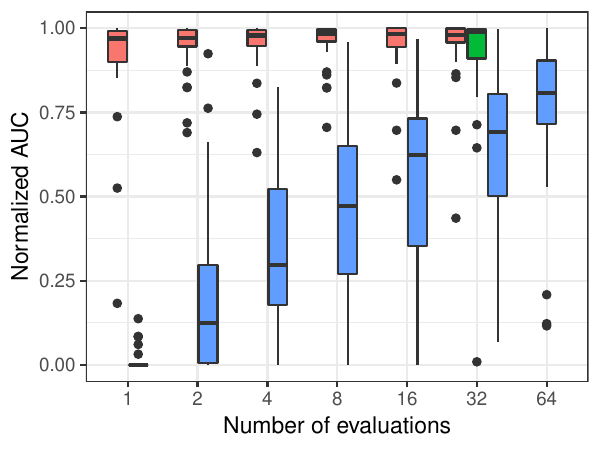}
     }
     \subfigure[Gradient boosting] {
       \includegraphics[width=.3\textwidth]{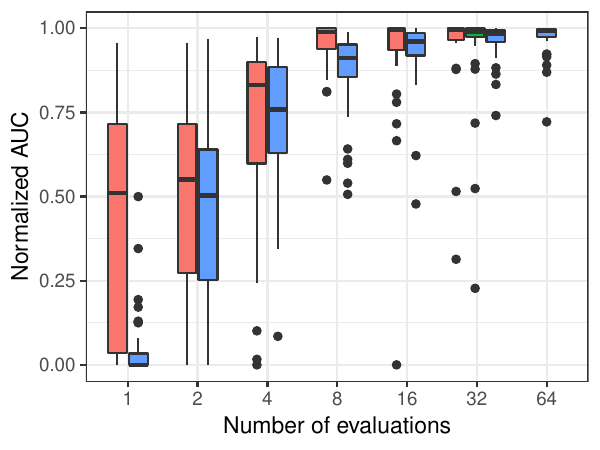}
     } \\
     \vspace{-0.1cm}
     \subfigure[Adaboost] {
       \includegraphics[width=.3\textwidth]{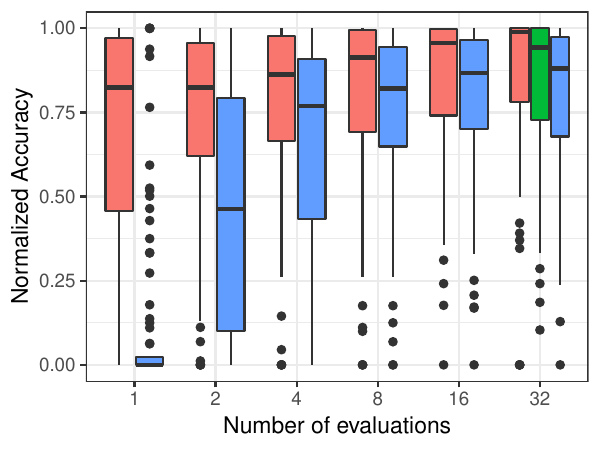}
     }
     \subfigure[Random forest] {
       \includegraphics[width=.3\textwidth]{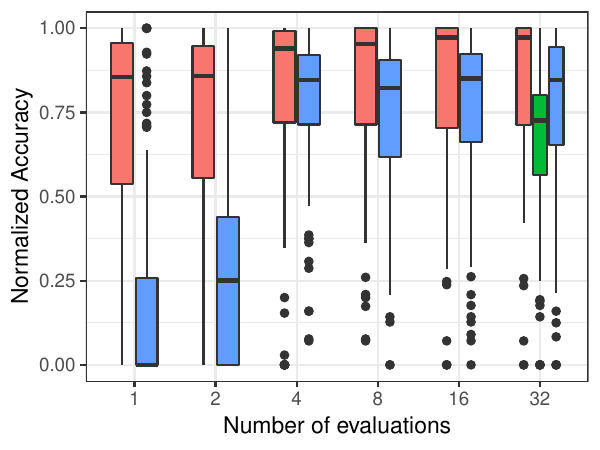}
     }
     \subfigure[SVM] {
       \includegraphics[width=.3\textwidth]{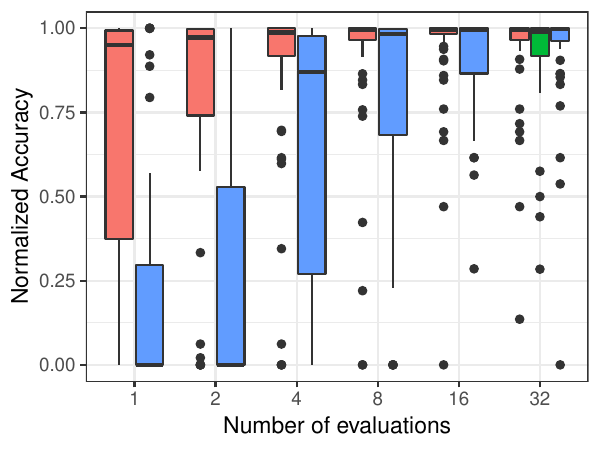}
     }
     \caption{Boxplots for different algorithms, comparing \textit{sets of defaults},
     \textit{random search} and \textit{Bayesian Optimization (mbo)} across several budgets. The y-axis depicts normalized Area under the Curve (upper) and normalized Accuracy (lower) across each learner and task.\label{fig:boxplots}}
   \end{center}
 \end{figure*}

\section{Exact Discretized vs. Greedy Defaults}
\label{sec:exp:exact}

We compare the computationally expensive exact discretized approach to the greedy approach in a small-scale experiment, to understand their relative performance in terms of hold-out accuracy.
In this experiment, we generate $n = \{1, 2, 3, 4, 5, 6\}$ defaults using both the greedy approach and the exact discretized  search, on a subset of the SVM hyperparameter space, for the $100$ datasets from the OpenML100. We aim to optimize the gamma ($[2^{-15}, 2^{3}]$, log-scale) and complexity ($[2^{-5}, 2^{15}]$, log-scale) hyperparameter for the RBF kernel, as \citeinline{Rijn2018} found these to be most important. We discretized the problem to have $16$ choices for both hyperparameters. Note that in order to obtain $6$ defaults, already $\binom{256}{6} \approx 3.7 \cdot 10^{11}$ possible options need to be evaluated, over $100$ datasets. 

Figure~\ref{fig:exact} shows the results. 
Note that by definition, the results for $1$ default should be approximately equal, and can only deviate when a tie is broken in a different way. 
Furthermore, even though intuitively the exact discretized approach should come up with better set defaults, this might not hold in practice, as the defaults are evaluated on datasets that were not considered when calculating the defaults. 
The results reveal that the sets of defaults from both strategies perform approximately the same.  As the greedy defaults have the benefit of being computationally much cheaper and provide anytime capabilities, we will use the greedy method for the remainder of the paper.

\section{Greedy Defaults} 
\label{sec:exp:greedy}
Figure~\ref{fig:boxplots} presents the results of the set of defaults obtained by the greedy approach and the baselines. Each subfigure shows the results for a given algorithm. The boxplots represent how the algorithm performed across the 38 (\texttt{mlR} algorithms) or 100 (\texttt{scikit-learn} algorithms) datasets that it was ran on. 
Results are normalized to $[0, 1]$ per algorithm and task using the best and worst result on each dataset across all settings.

\begin{figure*}[tb!]
   \begin{center}
     \subfigure[Elastic net] {
       \includegraphics[width=.32\textwidth]{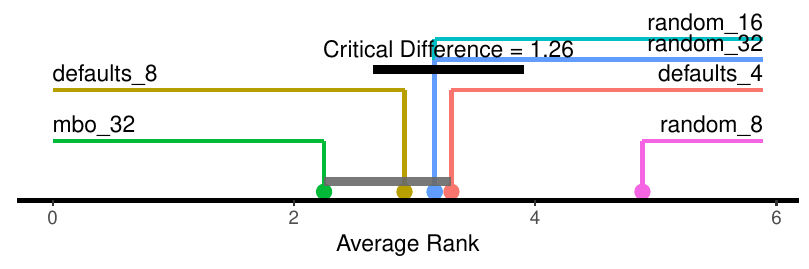}
     }
     \subfigure[Decision tree] {
       \includegraphics[width=.32\textwidth]{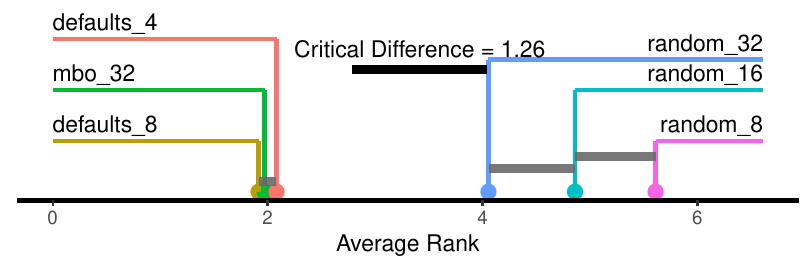}
     }
     \subfigure[Gradient boosting] {
       \includegraphics[width=.32\textwidth]{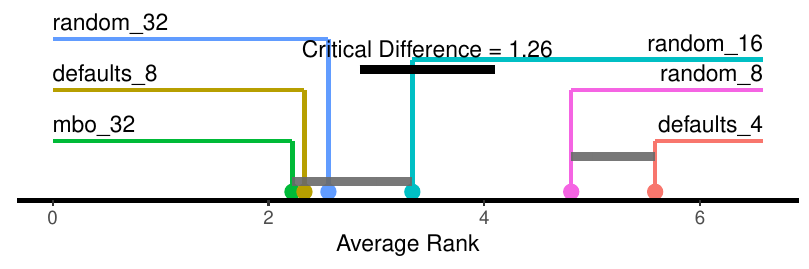}
     } \\
     \vspace{-0.1cm}
     \subfigure[Adaboost] {
       \includegraphics[width=.32\textwidth]{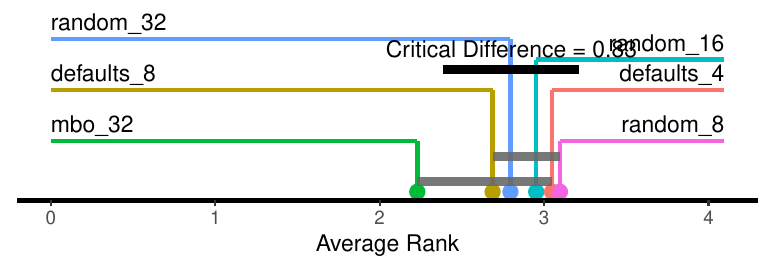}
     }
     \subfigure[Random forest] {
       \includegraphics[width=.32\textwidth]{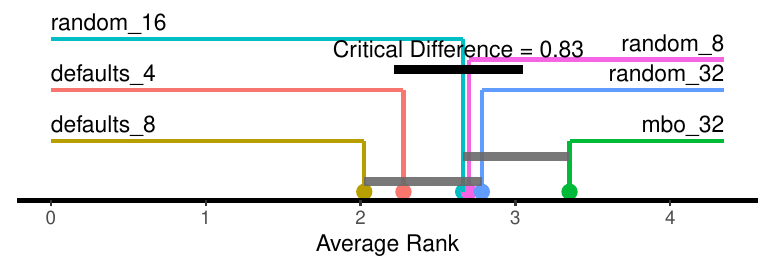}
     }
     \subfigure[SVM] {
       \includegraphics[width=.32\textwidth]{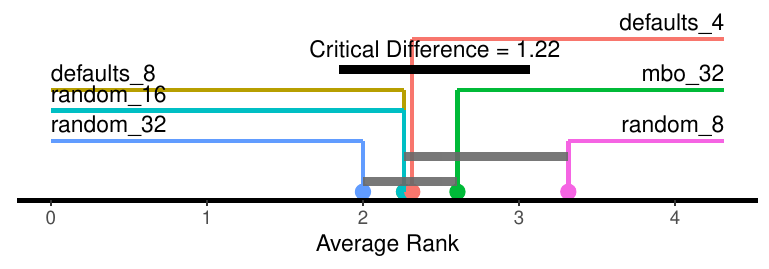}
     }
     \caption{Critical Differences plots comparing $4$ and $8$ defaults with random search with $(8, 16$ and $32)$ and Bayesian Optimization with $32$ iterations. The x-axis contains average ranks across all datasets for which all stategies terminated.\label{fig:cdplot}}
   \end{center}
 \end{figure*}

The results reveal some expected trends. For both the defaults and the random search, having more iterations strictly improves performance.
As might be expected, random search with only 1 or 2 iterations does not seem to be a compelling strategy. 
Bayesian Optimization is often among the highest ranked strategies (which can also be seen in Figure~\ref{fig:cdplot}).
We further observe that using only a few defaults is already competitive with Bayesian Optimization and random search strategies with higher budget. 
In many cases, the defaults are competitive with random search procedures that have four to eight times more budget. 
This is in particular clear for decision trees and elastic net, where $4$ defaults already outperform random search significantly.
In some cases, e.g., for random forest, advantages of using defaults over random search with multiples of the budgets seem negligible. A reason for this could be, that random forests in general appear more robust with regards to selection of hyperparameters~\cite{Probst2018}, and thus do not profit as much from optimal defaults. We can also see, that random search seems to stagnate much quicker than the set of defaults, which suggests that defaults can still be a viable alternative.
It can also be seen, that defaults perform particularly well when the budget is low. When the budget increases, the potential gains decrease. 
This can be observed in Figures \ref{fig:boxplots} d)-e), where performance only increases marginally after $8$ defaults.
This can be intuitively understood from the fact, that defaults are learned from a limited number of datasets. Thus when the number of defaults approaches the number of datasets, defaults more and more adapt to a small set of datasets, rather than generalizing to many datasets.

In order to further analyze the results, we perform the Friedman statistical test (with post-hoc Nemenyi test) on the results~\cite{Demsar2006}. 
Again, per classifier and task combination, each strategy gets assigned a rank. The ranks are averaged over all datasets, and reported in Figures \ref{fig:cdplot}.
If the difference is bigger than the critical distance, there is statistical proof that this difference in performance was not due to random chance. 
Each pair of strategies that is connected by a gray line, is considered statistically equivalent at $\alpha = 0.05$.
We observe the same trends as in the boxplots. Strategies that employ defaults are usually ranked better than random search with 2-4 times the budget, and significantly better then using the same budget.
Discrepancies between learners in \ref{fig:boxplots} a) - c) and d)-f) can stem for example from the fact that fewer experimental results were available for the latter, hampering the performance of trained surrogate models.

Numbers of datasets used in the different comparisons arise due to computational constraints. For the evaluation of elastic net, decision trees, the largest two datasets have been excluded from the evaluation, thus allowing for an evaluation on 36 datasets. For Adaboost, random forest and SVM, only datasets where all evaluations finished are included.
\section{Conclusions} 

% Sets of default parameters can significantly improve over random search or Bayesian Optimization if the number of available is comparably small. We propose a method for obtaining such sets of default configurations from large collections of experiment results, that are often already available, for example from \texttt{OpenML} \cite{Vanschoren2014}.
% Those defaults can be easily implemented in different, widely-used software packages or distributed as simple tables. Due to their simple nature, even inexperienced users can easily try different configurations in order to obtain better results. Especially in situations, where the number of available evaluations, for example due to time- or computational constraints, those defaults can be a valuable alternative.

We explored the potential of using sets of defaults. 
Single defaults usually give poor performance, whereas a more complex optimization procedure can be hard to implement and often needs many iterations. Instead, we show that using a sequence of defaults can be a robust and viable alternative to obtain a good hyperparameter configuration. 
When having access to large amounts of historic results, we can infer short lists of 4-8 candidate configurations that are competitive to random search or Bayesian Optimization with much larger budgets.

Finding the defaults in itself is an NP-hard search problem. We proposed a Mixed Integer Programming solution and a greedy solution, the latter running in polynomial time. Both strategies seem to obtain comparable results, which validates the use of the greedy strategy. An additional benefit of a greedy strategy is its \textit{anytime} performance, which allows selecting an arbitrarily sized, optimal subset.

We performed an extensive evaluation across 6 algorithms, 2 workbenches and 2 performance measures (accuracy and area under the curve). We compared the optimization over sets of defaults against random search with a a much larger budget, and Bayesian Optimization. 
Experiments showed that defaults consistently outperform the other strategies, even when given less budget. 

We note that using sets of defaults is especially worthwhile when either computation time or expertise on hyperparameter optimization is lacking. Especially in the regime of few function evaluations, sets of defaults seem to work particularly well and are statistically equivalent with state-of-the-art techniques. A potential drawback of our method is that our defaults are optimal with respect to a single metric such as accuracy or AUC, and thus might need to be used separately for different evaluation metrics. Identifying whether is actually the case requires further investigation.

However, when fixing the metric, our results can readily be implemented in machine learning software as simple, hard-coded lists of parameters. These will require less knowledge of hyperparameter optimization from the users than current methods, and lead to faster results in many cases.

In future work, we aim to incorporate multi-objective measures for determining the defaults. This includes, but is not limited to computational time, as focusing on defaults that are expected to run fast, might improve the anytime performance even further.
%Additionally, we aim to combine search spaces from various algorithms into one set of defaults.

In order to improve performance even more, models trained for a set of defaults can be ensembled. 
Finally, we aim to combine the sets of defaults with other hyperparameter optimization techniques, e.g., by using it to warm-start Bayesian Optimization procedures. 
Successfully combining these two paradigms might be the key to convincingly push forward the state-of-the-art for Automated Machine Learning.

~\\\noindent{}\textbf{Acknowledgements.}
We would like to thank Michael Hennebry for the discussion leading to the formulation of the `exact discretized' approach. 
Furthermore, we would like to thank Fabian Scheipl for constructive criticism of the manuscript. 
This work has been funded by the German Federal Ministry of Education and Research (BMBF) under Grant No. 01IS18036A. 
The authors of this work take full responsibilities for its content.

\bibliography{multipledefaults}

\begin{thebibliography}{}

\bibitem[\protect\citeauthoryear{Bergstra and Bengio}{2012}]{Bergstra2012}
Bergstra, J., and Bengio, Y.
\newblock 2012.
\newblock Random search for hyper-parameter optimization.
\newblock {\em JMLR} 13(Feb):281--305.

\bibitem[\protect\citeauthoryear{Bergstra \bgroup et al\mbox.\egroup
  }{2011}]{Bergstra2011}
Bergstra, J.; Bardenet, R.; Bengio, Y.; and K\'{e}gl, B.
\newblock 2011.
\newblock Algorithms for hyper-parameter optimization.
\newblock In {\em Proc. of NIPS}. Curran Associates, Inc.
\newblock  2546--2554.

\bibitem[\protect\citeauthoryear{Bischl \bgroup et al\mbox.\egroup
  }{2016}]{Bischl2016}
Bischl, B.; Lang, M.; Kotthoff, L.; Schiffner, J.; Richter, J.; Studerus, E.;
  Casalicchio, G.; and Jones, Z.~M.
\newblock 2016.
\newblock mlr: Machine learning in {R}.
\newblock {\em JMLR} 17(170):1--5.

\bibitem[\protect\citeauthoryear{Bischl \bgroup et al\mbox.\egroup
  }{2017}]{Bischl2017}
Bischl, B.; Casalicchio, G.; Feurer, M.; Hutter, F.; Lang, M.; Mantovani,
  R.~G.; van Rijn, J.~N.; and Vanschoren, J.
\newblock 2017.
\newblock {OpenML Benchmarking Suites and the OpenML100}.
\newblock {\em arXiv preprint arXiv:1708.03731v1}.

\bibitem[\protect\citeauthoryear{Bischl \bgroup et al\mbox.\egroup
  }{2018}]{mlrMBO}
Bischl, B.; Richter, J.; Bossek, J.; Horn, D.; Thomas, J.; and Lang, M.
\newblock 2018.
\newblock {{mlrMBO}}: {{A Modular Framework}} for {{Model}}-{{Based
  Optimization}} of {{Expensive Black}}-{{Box Functions}}.
\newblock {\em arXiv preprint arXiv:1703.03373}.

\bibitem[\protect\citeauthoryear{Brazdil \bgroup et al\mbox.\egroup
  }{2008}]{Brazdil2008}
Brazdil, P.; Giraud-Carrier, C.; Soares, C.; and Vilalta, R.
\newblock 2008.
\newblock {\em Metalearning: Applications to Data Mining}.
\newblock Springer Publishing Company, Incorporated, 1 edition.

\bibitem[\protect\citeauthoryear{Cawley and Talbot}{2010}]{cawley10}
Cawley, G.~C., and Talbot, N.~L.
\newblock 2010.
\newblock On over-fitting in model selection and subsequent selection bias in
  performance evaluation.
\newblock {\em J. Mach. Learn. Res.} 11:2079--2107.

\bibitem[\protect\citeauthoryear{Chen and Guestrin}{2016}]{xgboost}
Chen, T., and Guestrin, C.
\newblock 2016.
\newblock {XGBoost}: A scalable tree boosting system.
\newblock In {\em Proc. of KDD}, KDD '16,  785--794.
\newblock New York, NY, USA: ACM.

\bibitem[\protect\citeauthoryear{Dem{\v{s}}ar}{2006}]{Demsar2006}
Dem{\v{s}}ar, J.
\newblock 2006.
\newblock {Statistical Comparisons of Classifiers over Multiple Data Sets}.
\newblock {\em JMLR} 7:1--30.

\bibitem[\protect\citeauthoryear{Eggensperger \bgroup et al\mbox.\egroup
  }{2015}]{Eggensperger2015}
Eggensperger, K.; Hutter, F.; Hoos, H.; and Leyton-Brown, K.
\newblock 2015.
\newblock Efficient benchmarking of hyperparameter optimizers via surrogates.
\newblock In {\em Proc. AAAI},  1114--1120.

\bibitem[\protect\citeauthoryear{Feurer \bgroup et al\mbox.\egroup
  }{2015}]{Feurer2015}
Feurer, M.; Klein, A.; Eggensperger, K.; Springenberg, J.~T.; Blum, M.; and
  Hutter, F.
\newblock 2015.
\newblock Efficient and robust automated machine learning.
\newblock In {\em Proc. of NIPS}. Curran Associates, Inc.
\newblock  2962--2970.

\bibitem[\protect\citeauthoryear{Feurer \bgroup et al\mbox.\egroup
  }{2018}]{Feurer2018}
Feurer, M.; Eggensperger, K.; Falkner, S.; Lindauer, M.; and Hutter, F.
\newblock 2018.
\newblock Practical automated machine learning for the automl challenge 2018.
\newblock In {\em ICML 2018 AutoML Workshop}.

\bibitem[\protect\citeauthoryear{Feurer, Springenberg, and
  Hutter}{2015}]{Feurer2015b}
Feurer, M.; Springenberg, J.~T.; and Hutter, F.
\newblock 2015.
\newblock Initializing bayesian hyperparameter optimization via meta-learning.
\newblock In {\em Proc. AAAI},  1128--1135.
\newblock AAAI Press.

\bibitem[\protect\citeauthoryear{Friedman, Hastie, and
  Tibshirani}{2010}]{glmnet}
Friedman, J.; Hastie, T.; and Tibshirani, R.
\newblock 2010.
\newblock Regularization paths for generalized linear models via coordinate
  descent.
\newblock {\em Journal of Statistical Software} 33(1):1--22.

\bibitem[\protect\citeauthoryear{Gomes \bgroup et al\mbox.\egroup
  }{2012}]{Gomes2012}
Gomes, T. A.~F.; Prud{\^e}ncio, R. B.~C.; Soares, C.; Rossi, A. L.~D.; and
  Carvalho, A.
\newblock 2012.
\newblock Combining meta-learning and search techniques to select parameters
  for support vector machines.
\newblock {\em Neurocomputing} 75(1):3--13.

\bibitem[\protect\citeauthoryear{Hall \bgroup et al\mbox.\egroup
  }{2009}]{Hall2009}
Hall, M.; Frank, E.; Holmes, G.; Pfahringer, B.; Reutemann, P.; and Witten,
  I.~H.
\newblock 2009.
\newblock {The WEKA Data Mining Software: An Update}.
\newblock {\em ACM SIGKDD explorations newsletter} 11(1):10--18.

\bibitem[\protect\citeauthoryear{Hodges and Lehmann}{1952}]{hodges1952}
Hodges, J.~L., and Lehmann, E.~L.
\newblock 1952.
\newblock The use of previous experience in reaching statistical decisions.
\newblock {\em Ann. Math. Statist.} 23(3):396--407.

\bibitem[\protect\citeauthoryear{Hutter, Hoos, and
  Leyton-Brown}{2011}]{Hutter2011}
Hutter, F.; Hoos, H.~H.; and Leyton-Brown, K.
\newblock 2011.
\newblock Sequential model-based optimization for general algorithm
  configuration.
\newblock In {\em Proc. of LION},  507--523.
\newblock Springer.

\bibitem[\protect\citeauthoryear{Jamieson and Talwalkar}{2016}]{Jamieson2016}
Jamieson, K., and Talwalkar, A.
\newblock 2016.
\newblock Non-stochastic best arm identification and hyperparameter
  optimization.
\newblock In {\em Proc. of AISTATS}, volume~51,  240--248.
\newblock PMLR.

\bibitem[\protect\citeauthoryear{Lavesson and Davidsson}{2006}]{Lavesson2006}
Lavesson, N., and Davidsson, P.
\newblock 2006.
\newblock Quantifying the impact of learning algorithm parameter tuning.
\newblock In {\em Proc. of AAAI}, volume~6,  395--400.

\bibitem[\protect\citeauthoryear{Leite, Brazdil, and
  Vanschoren}{2012}]{Leite2012}
Leite, R.; Brazdil, P.; and Vanschoren, J.
\newblock 2012.
\newblock {Selecting Classification Algorithms with Active Testing}.
\newblock In {\em Proc. of MLDMPR}. Springer.
\newblock  117--131.

\bibitem[\protect\citeauthoryear{Li \bgroup et al\mbox.\egroup }{2017}]{Li2017}
Li, L.; Jamieson, K.; DeSalvo, G.; Rostamizadeh, A.; and Talwalkar, A.
\newblock 2017.
\newblock Hyperband: {Bandit}-{Based} {Configuration} {Evaluation} for
  {Hyperparameter} {Optimization}.
\newblock In {\em Proc.~of ICLR 2017}.

\bibitem[\protect\citeauthoryear{Nemhauser, Wolsey, and
  Fisher}{1978}]{Nemhauser1978}
Nemhauser, G.~L.; Wolsey, L.~A.; and Fisher, M.~L.
\newblock 1978.
\newblock An analysis of approximations for maximizing submodular set
  functions—i.
\newblock {\em Mathematical programming} 14(1):265--294.

\bibitem[\protect\citeauthoryear{Pedregosa \bgroup et al\mbox.\egroup
  }{2011}]{Pedregosa2011}
Pedregosa, F.; Varoquaux, G.; Gramfort, A.; Michel, V.; Thirion, B.; Grisel,
  O.; Blondel, M.; Prettenhofer, P.; Weiss, R.; Dubourg, V.; Vanderplas, J.;
  Passos, A.; Cournapeau, D.; Brucher, M.; Perrot, M.; and Duchesnay, E.
\newblock 2011.
\newblock Scikit-learn: Machine learning in {P}ython.
\newblock {\em JMLR} 12:2825--2830.

\bibitem[\protect\citeauthoryear{Pinto, Soares, and
  Mendes-Moreira}{2016}]{Pinto2016}
Pinto, F.; Soares, C.; and Mendes-Moreira, J.
\newblock 2016.
\newblock Towards automatic generation of metafeatures.
\newblock In {\em Proc. of PAKDD},  215--226.
\newblock Springer.

\bibitem[\protect\citeauthoryear{Probst, Boulesteix, and
  Bischl}{2019}]{Probst2018}
Probst, P.; Boulesteix, A.-L.; and Bischl, B.
\newblock 2019.
\newblock Tunability: Importance of hyperparameters of machine learning
  algorithms.
\newblock {\em JLMR} 20(53):1--32.

\bibitem[\protect\citeauthoryear{Snoek, Larochelle, and
  Adams}{2012}]{Snoek2012}
Snoek, J.; Larochelle, H.; and Adams, R.~P.
\newblock 2012.
\newblock Practical bayesian optimization of machine learning algorithms.
\newblock In {\em Proc. of NIPS}, NIPS'12,  2951--2959.
\newblock USA: Curran Associates Inc.

\bibitem[\protect\citeauthoryear{Therneau and Atkinson}{2018}]{rpart}
Therneau, T., and Atkinson, B.
\newblock 2018.
\newblock {\em rpart: Recursive Partitioning and Regression Trees}.
\newblock R package version 4.1-13.

\bibitem[\protect\citeauthoryear{van Rijn and Hutter}{2018}]{Rijn2018}
van Rijn, J.~N., and Hutter, F.
\newblock 2018.
\newblock Hyperparameter importance across datasets.
\newblock In {\em Proc. of KDD},  2367--2376.
\newblock ACM.

\bibitem[\protect\citeauthoryear{van Rijn}{2016}]{Rijn2016}
van Rijn, J.~N.
\newblock 2016.
\newblock {\em Massively Collaborative Machine Learning}.
\newblock Ph.D. Dissertation, Leiden University.

\bibitem[\protect\citeauthoryear{Vanschoren \bgroup et al\mbox.\egroup
  }{2014}]{Vanschoren2014}
Vanschoren, J.; van Rijn, J.~N.; Bischl, B.; and Torgo, L.
\newblock 2014.
\newblock {OpenML: networked science in machine learning}.
\newblock {\em ACM SIGKDD Explorations Newsletter} 15(2):49--60.

\bibitem[\protect\citeauthoryear{Wistuba, Schilling, and
  Schmidt-Thieme}{2015a}]{Wistuba2015learning}
Wistuba, M.; Schilling, N.; and Schmidt-Thieme, L.
\newblock 2015a.
\newblock Learning hyperparameter optimization initializations.
\newblock In {\em Proc. of DSAA},  1--10.
\newblock IEEE.

\bibitem[\protect\citeauthoryear{Wistuba, Schilling, and
  Schmidt-Thieme}{2015b}]{wistuba15b}
Wistuba, M.; Schilling, N.; and Schmidt-Thieme, L.
\newblock 2015b.
\newblock Sequential model-free hyperparameter tuning.
\newblock In {\em Proc. of ICDM},  1033--1038.

\bibitem[\protect\citeauthoryear{Yogatama and Mann}{2014}]{yogatama14}
Yogatama, D., and Mann, G.
\newblock 2014.
\newblock {Efficient Transfer Learning Method for Automatic Hyperparameter
  Tuning}.
\newblock In {\em Proc. of AISTATS}, volume~33,  1077--1085.
\newblock PMLR.

\end{thebibliography}
\bibliographystyle{aaai}

\end{document}